\documentclass[conference]{IEEEtran}

\IEEEoverridecommandlockouts
\usepackage{cite}
\usepackage{amsmath,amssymb,amsfonts}
\usepackage{algorithmic}
\usepackage[ruled,vlined]{algorithm2e}
\usepackage{comment}
\usepackage{graphicx}
\usepackage{microtype}
\usepackage{caption}
\usepackage{subcaption}
\usepackage{textcomp}
\usepackage{xcolor}
\usepackage{booktabs}
\usepackage{multirow}
\usepackage{xcolor}

\def\BibTeX{{\rm B\kern-.05em{\sc i\kern-.025em b}\kern-.08em
    T\kern-.1667em\lower.7ex\hbox{E}\kern-.125emX}}
    
\newcommand{\cachesample}{\textit{ES-SpMM}}
\newcommand{\tit}[1]{\textit{#1}}

\newcommand{\buc}{\textit{Bucket}}
\newcommand{\fr}{\textit{FastRand}}
\newcommand{\liang}[1]{\textcolor{red}{(LiangLuo: #1)}}
\newcommand{\chien}[1]{\textcolor{blue}{(ChienYu: #1)}}

\begin{document}

\title{Accelerating SpMM Kernel with Cache-First Edge Sampling for Graph Neural Networks 
\thanks{
}
}

\author{\IEEEauthorblockN{
Chien-Yu Lin}
\IEEEauthorblockA{
Univeristy of Washington \\
\textit{Seattle, WA, USA} \\
cyulin@cs.washington.edu
}
\and
\IEEEauthorblockN{
Liang Luo}
\IEEEauthorblockA{
Facebook Inc. \\
\textit{Seattle, WA, USA} \\
liangluo@fb.com}
\and
\IEEEauthorblockN{
Luis Ceze}
\IEEEauthorblockA{
University of Washington, OtcoML \\
\textit{Seattle, WA, USA} \\
luisceze@cs.washington.edu}
}

\maketitle

\begin{abstract}
Graph neural networks (GNNs), an emerging deep learning model class, can extract meaningful representations from highly expressive graph-structured data and are therefore gaining popularity for wider ranges of applications.
However, current GNNs suffer from the poor performance of their sparse-dense matrix multiplication (SpMM) operator, even when using powerful GPUs.
Our analysis shows that 95\% of the inference time could be spent on SpMM when running popular GNN models on NVIDIA's advanced V100 GPU.
Such SpMM performance bottleneck hinders GNNs' applicability to large-scale problems or the development of more sophisticated GNN models.

To address this inference time bottleneck, we introduce \cachesample{}, a cache-first edge sampling mechanism and codesigned SpMM kernel.
\cachesample{} uses edge sampling to downsize the graph to fit into GPU's shared memory.
It thus reduces the computation cost and improves SpMM's cache locality.
To evaluate \cachesample's performance, we integrated it with a popular GNN framework, DGL, and tested it using representative GNN models and datasets.
Our results show that \cachesample{} outperforms the highly optimized \tit{cuSPARSE} SpMM kernel by up to 4.35x with no accuracy loss and by 45.3x with less than a 1\% accuracy loss.

\end{abstract}

\section{Introduction}
\label{sec:intro}
GNNs are a class of  powerful deep learning (DL) models that can extract high-level embeddings from graph-structured data. Because the graph structure can represent many different types of information, GNN has a wide-range of applications, including network data mining \cite{gcn, graphsage}, program synthesis \cite{gnnapp_program}, physics modeling \cite{gnnapp_physics} and medical decision making \cite{gnnapp_drug}. As a result, this model is attracting significant attention from both academia and industry. 

\begin{figure}[t!]
    \centering
    \includegraphics[width=1\linewidth]{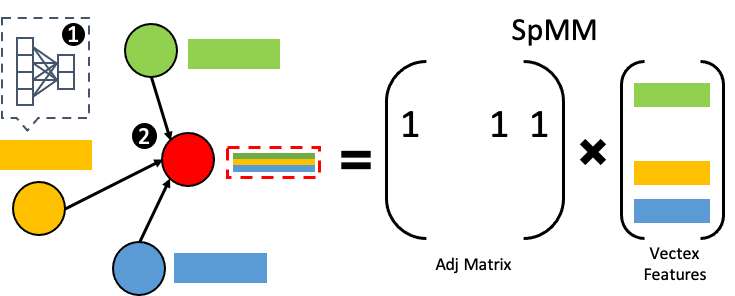}
    \caption{
    Basic operations of a typical GNN layer (left). Given graph data with vertices, edges, and vertex features, a GNN layer: (1) feeds the features to an MLP (feature extraction) using the GEMM operation; (2) sums the features from the neighborhood (feature aggregation) using an SpMM operation (right).}
    \label{fig:gnn_spmm}
\end{figure}

Unfortunately, GNNs are notoriously inefficient to run because of their sparse operations \cite{featgraph, gespmm}.
As Figure \ref{fig:gnn_spmm} shows, a typical GNN layer uses a multi-layer perceptron (MLP) to extract dense node features and aggregate the features according to the edges of the graph.
In practice, MLP feature extraction uses the general matrix multiplication (GEMM) operation, and feature aggregation uses the sparse-dense matrix multiplication (SpMM) or an SpMM-like operation.
Since GNNs usually run on GPUs, where the hardware architecture and memory system are optimized for dense matrix operations, the GPU time spent on the dense GEMM operation is relatively short, hence the performance bottleneck usually lies in the sparse SpMM operation.
In our analysis, for a standard GCN model \cite{gcn}, the SpMM kernel can consume up to 95\% of the end-to-end inference time.

To make SpMM run faster on GPUs, researchers have explored different SpMM kernel optimizations, including
leveraging GPU's Instruction-Level Parallelism (ILP) \cite{merge-spmm}, Thread-Level Parallelism (TLP) \cite{merge-spmm}, and cache memory \cite{merge-spmm, gespmm}; improving work balance; and using customized sparse formats \cite{ellpack_spmm, block_spmm}. 
However, these methods still fall short of speeding GNN inference performance on GPUs as previous literatures only show limited speedup (10\% to 30\%) \cite{merge-spmm, gespmm, aspt} over the standard \tit{cuSPARSE} kernel or is optimized for problems with lower sparsity, e.g., pruned neural networks \cite{gale2020sparse}.
It is thus critical to broader solutions than kernel optimizations alone to meet this goal.

Independently, edge sampling techniques \cite{DropEdge, bayesian_edge_sample, neural_sparse} proposed by researchers allow the training of possibly deeper and more accurate GNN models by solving the over-smoothing and over-fitting problems \cite{li2018, jknet, klicpera2019}. 
Since edge sampling reduces the number of connections in the graph, which in turn reduces SpMM computation time, it could potentially be used to speed up the SpMM kernel. 
However, existing methods requires preprocessing the graph on the CPU \cite{DropEdge} or using a complex model to sample the edges \cite{neural_sparse}. 
The resulting added overheads eventually eclipse the benefit of reduced computational complexity. 

To more efficiently leverage edge sampling for GNN acceleration, we introduce \cachesample{}, a sampling algorithm and codesigned SpMM kernel to speed GNN inference performance. 
Key \cachesample{} concepts include:
(1) an \textit{in-kernel sampling mechanism}, which eliminates preprocessing overheads and leverages a GPU's parallel computing power to accelerate the sampling process,
(2) a \textit{cache-first edge sampling design} that fits the entire sampled graph on a GPU's shared memory by using the shared memory size as the sampling target, reducing compute load while improving cache locality,
(3) \textit{coalesced global memory access} to efficiently fetch the sampled graph data and load shared memory.
(4) and two \textit{lightweight sampling strategies} to use in conjunction with the kernel to achieve optimal speedup with minimal accuracy.
The two strategies offer different trade-offs between model accuracy and speedup and are adaptable to the dataset specifics.
Using these novel optimizations and designs, \cachesample{} significantly outperforms state-of-the-art SpMM implementations with negligible accuracy loss for GNNs' workloads. 

To verify the effectiveness of our design, we integrated \cachesample{} into DGL, a popular graph learning library that provides state-of-the-art GNN runtime performance.
We designed \cachesample{} to work with the standard CSR (Compressed Sparse Row) format to avoid format transformation overhead, which lets us seamlessly swap DGL's backend SpMM kernel with \cachesample{} with no modification to user-level code. 
While the proposed techniques in \cachesample{} are sufficiently general to benefit both GNN training and inference, we focus here on inference only.
We conducted comprehensive experiments using two popular GNN models, GCN and GraphSage, on four representative graph benchmarks.
Our evaluation results for inference time show that \cachesample{} outperformed \tit{cuSPARSE} by up to 4.35x with no accuracy loss and 45.3x with less than a 1\% accuracy loss.

In sum, our main contributions include:
\begin{itemize}
  \item The design and implementation of \cachesample{}, a cache-first edge sampling mechanism and SpMM kernel codesign, to speed up GNN inference.
  \item A novel in-kernel edge sampling mechanism that accelerates the sampling process and eliminates the preprocessing overhead. 
  \item Two lightweight sampling strategies to achieve optimal speedup with minimal accuracy loss across different datasets. 
  \item Standard CSR format support and the integration with a popular GNN framework, DGL, without touching user-level code.
  \item Comprehensive evaluations on representative GNN models and datasets to verify \cachesample{}'s significantly improved inference efficiency.
  \item To the best of our knowledge, \cachesample{} is the first work that uses edge sammpling to accelerate GNN inference.
\end{itemize}

\section{Background and Motivation}
\label{sec:background}

In this section, we now briefly describe the GNN, the Compressed Sparse Row (CSR) format, GPU preliminaries, relevant SpMM optimization on GPUs, and the current state of the GNN computational bottleneck.

\subsection{Graph Neural Networks}
GNNs target feature-rich graphs with have high-dimensional, dense features associated with nodes or edges. 
A typical GNN has a multi-layer structure. 
Within a layer, it uses an MLP to extract high-level features from the input features, aggregate the extracted features based on the graph structure and apply an activation function like ReLU on the output features (see Figure \ref{fig:gnn_spmm}).
Activated features become the input features for the next layer. 
The final layer's output features are then be used for downstream prediction tasks, such as node classification or edge prediction.

GNNs differ from basic neural networks mainly due to feature aggregation.
Among the different feature aggregation methods, \tit{Sum} is the most commonly used feature aggregator. 
In \tit{Sum} aggregation, a node's new features are the sum of its neighborhood nodes' features.
Equation \ref{eq:gcn} shows the formalized computation of a GNN layer using \tit{Sum} aggregation.
$H^{(l)}$ and $W^{(l)}$ are node features and MLP weights for layer $l$, $A$ is the adjacency (adj) matrix, $\sigma (\cdot)$ is the activation function, and the output of the equation becomes the next layer's node features ($H^{(l+1)}$).

\begin{equation} \label{eq:gcn}
H^{(l+1)} = \sigma ( A H^{(l)} W^{(l)})
\end{equation}

Here, the computation of $H^{(l)} W^{(l)}$ is essentially a GEMM since both $H$ and $W$ are dense matrices. 
The multiplication of $A$ and $H^{(l)} W^{(l)}$ is an SpMM operation because $A$ is a sparse matrix and the result of $H^{(l)} W^{(l)}$ is a dense matrix. 
Since \tit{Sum} is an effective aggregator and adopted by many GNN models \cite{gcn, graphsage, fastgcn, cluster-gcn, jknet, as-gcn, graphsaint, deepgcns, gcnii}, in this work we focus on accelerating the GNNs using \tit{Sum} aggregation.

\begin{figure}[t!]
    \centering
    \includegraphics[width=0.85\linewidth]{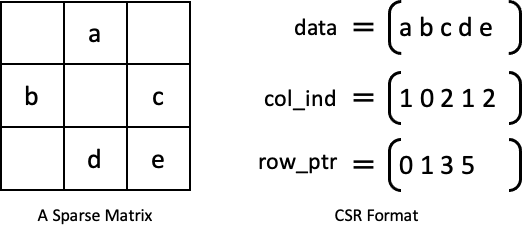}
    \caption{An example of the CSR sparse format. CSR represents a sparse matrix with three arrays: \tit{data}, \tit{col\_ind} (column indices), and \tit{row\_ptr} (row pointers). For an unweighted graph's adjacency matrix, $data$ is an array of 1.}
    \label{fig:csr}
\end{figure}

\subsection{Compressed-Sparse Row (CSR) Format}
\label{subsec:csr}
Sparse formats efficiently store sparse data; in the context of GNN, the sparse adjacency matrix is stored in a sparse format, which skips all zero values to compress the matrix. 

There are several common sparse formats. Here, we introduce the most standard one, CSR (see Figure \ref{fig:csr}).
CSR has three arrays to represent a 2D sparse matrix. 
The first, the \tit{data} array, keeps all non-zero values following the row-major order. 
For an unweighted graph's adjacency matrix, non-zero values are all ones, which is a common case in GNNs' workloads.
The second, \tit{col\_ind} (column indices), keeps column indices of each non-zero value and uses the same order as the \tit{data} array. 
The third, \tit{row\_ptr} (row pointers) array, records the accumulated number of non-zero (NNZ) values for each row.

Due to its high efficiency and simplicity, CSR has been adopted by virtually all of the platforms that support sparse operations. 
Further, most graph matrices come in the CSR format. 
Therefore, we design \cachesample{} to directly work with this format. 
By supporting CSR, \cachesample{} can serve as a drop-in replacement for any existing sparse system without the need to transform formats and incurring resulting large overheads.

\begin{table}[tb!]
\caption{Statistics of the evaluated graph datasets.}
\label{tab:dataset}
\begin{center}
\begin{small}
\begin{sc}
\scalebox{1}{
\begin{tabular}{lrrcr}
\toprule
Dataset & \# Node & \# Edge & Avg. Degree & \# Class \\
\midrule
Pubmed          & 19.7K     & 88.6K     & 4.5  & 3     \\
Arxiv      & 169K      & 2.3M      & 13.7  & 40    \\
Proteins   & 132K    & 79.1M     & 597   & 112   \\
Reddit          & 233K    & 115M    & 493   & 41    \\
\bottomrule
\end{tabular}}
\end{sc}
\end{small}
\end{center}
\end{table}

\subsection{GPU Preliminaries}
\label{subsec:gpu}
We next briefly introduce the GPU hardware architecture and execution model, using NVIDIA's terminology since this work is implemented in CUDA.

In terms of hardware architecture, a GPU is composed of an array of streaming multiprocessors (SMs).
Each SM consists of a few blocks of 32 CUDA cores, where a CUDA core is the smallest compute unit on the GPU.
It also has a piece of on-chip, software-managed shared memory (L1 cache) and a pool of registers. 
Beyond the SM level, there is an L2 cache and a large off-chip global memory shared by all SMs. 

For the execution model, a GPU kernel is typically executed by thousands of parallel threads, grouped into thread blocks; each thread block is mapped to an SM.
A thread block is further divided into warps, where each warp consists 32 threads.
A warp of threads is mapped to a block of 32 CUDA cores and executes the same instruction simultaneously; thus, this execution model is called Single Instruction Multiple Threads (SIMT).
The warps in a thread block are streamed into the SM by a hardware scheduler.
When a warp is idle, e.g., waiting for memory requests, the warp is context switched out in favor of other warps and is continued later.
Therefore, given sufficient warps in a thread block, it's possible to hide the memory latency with this execution model.
Furthermore, a GPU tries to combine memory requests from a warp into as few global memory transactions as possible.

\begin{figure}[t!]
    \centering
    \includegraphics[width=0.95\linewidth]{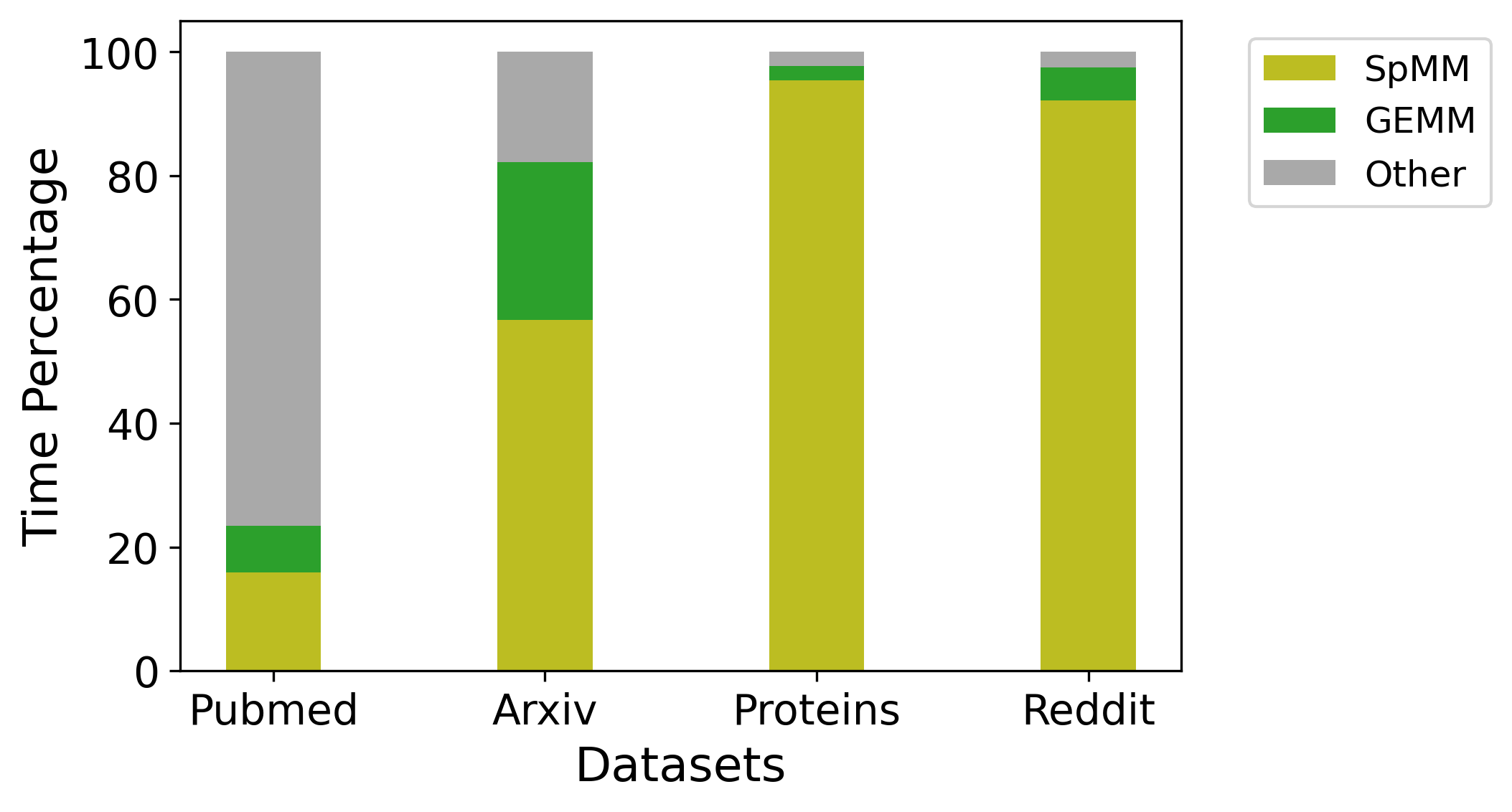}
    \caption{End-to-end compute time breakdown of the GCN model on different datasets.}
    \label{fig:breakdown}
\end{figure}

\subsection{Performance Bottleneck of GNNs}
\label{subsec:bottleneck}
To illustrate the performance bottleneck of GNNs, Figure \ref{fig:breakdown} breaks down the computation time of GCN \cite{gcn}, a popular GNN model, for four different graph datasets. 
The GCN model is implemented in DGL, which uses the cuBLAS and cuSPARSE libraries for the GEMM and SpMM operations, respectively. 
The experiments are conducted on an AWS p3.2xlarge instance with a NVIDIA's V100 GPU.
The statistics of the evaluated datasets and the model setup are shown on Table \ref{tab:dataset} and Table \ref{tab:model}.
The detailed descriptions of the datasets and models are in Section \ref{subsec:exp_setup}.

As Figure \ref{fig:breakdown} shows, despite the use of the highly optimized cuSPARSE library, the performance bottleneck of GNNs still lies in the SpMM kernel. 
For the smaller datasets, such as Pubmed and Arxiv, the SpMM kernel takes 16\% and 56\% of the end-to-end compute time, respectively. 
For the larger datasets, such as Proteins and Reddit, the SpMM kernel consumes an unacceptale 95.4\% of the total inference time.

These dramatic results are due to GPUs being optimized for dense and regular operations. 
SpMM's sparse and irregular features make it difficult to achieve high throughput on a GPU.
As the scale of the graphs involved in GNNs' workloads becomes progressively larger \cite{ogb}, the compute bottleneck on the SpMM kernel becomes increasingly more severe and hinders the development of larger, deeper GNN models.

To relieve the compute bottleneck and further accelerate GNN inference, we propose a novel SpMM kernel that we codesigned with an edge sampling mechanism to achieve an order of magnitude speedup over existing SpMM kernels.

\section{Edge Sampling for GNN's Inference}
\label{sec:edge_sample}
To demonstrate how we leverage edge sampling to accelerate SpMM kernel performance, in this section we briefly introduce edge sampling techniques in the context of GNNs, analyze its impacts on inference accuracy, and examine the limitations of existing methods.




\subsection{Edge Sampling}
\begin{figure}[t!]
    \centering
    \includegraphics[width=\linewidth]{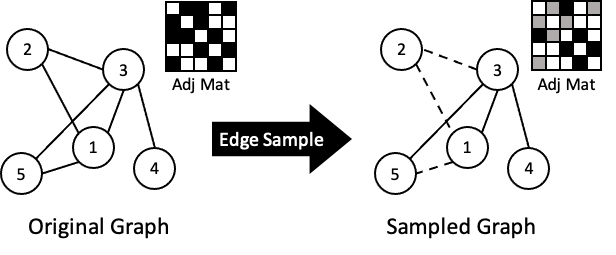}
    \caption{The edge sampling process. The NNZ of the adjacency matrix for the sampled graph is reduced compared to the original graph. 
    }
    \label{fig:edge_sample}
\end{figure}

Edge sampling is a technique to include only a subset of edges of the original graph while discarding the rest. The existing methods include randomly dropping edges \cite{DropEdge, bayesian_edge_sample} or using a learned neural network to determine which edges to abandon \cite{neural_sparse}. 
Figure \ref{fig:edge_sample} highlights the edge sampling process.

To date, edge sampling has been used to mitigate GNN's over-fitting and over-smoothing \cite{li2018, jknet, klicpera2019} training problems. 
For example, DropEdge \cite{DropEdge} randomly samples a different subset of edges throughout training iterations with a fixed dropping rate.
The results of DropEdge shows that such dynamic random edge sampling improves training accuracy for both shallow and deep GNN models.

However, rather than improving training accuracy, we for the first time use edge sampling to accelerate GNN's inference.


\subsection{Inference Accuracy}
\label{subsec:edge_sample_acc}
Unlike previous edge sampling works for GNN training have training loops and backpropagation to adapt the loss of edges, model accuracy will likely diminish with the edge-sampled graph during inference, as some graph information is lost in the process of sampling.
To understand whether edge sampling is applicable to GNN inference with an acceptable accuracy loss, we conducted an analysis of its effect on inference accuracy.

\begin{figure}[t!]
    \centering
    \includegraphics[width=\linewidth]{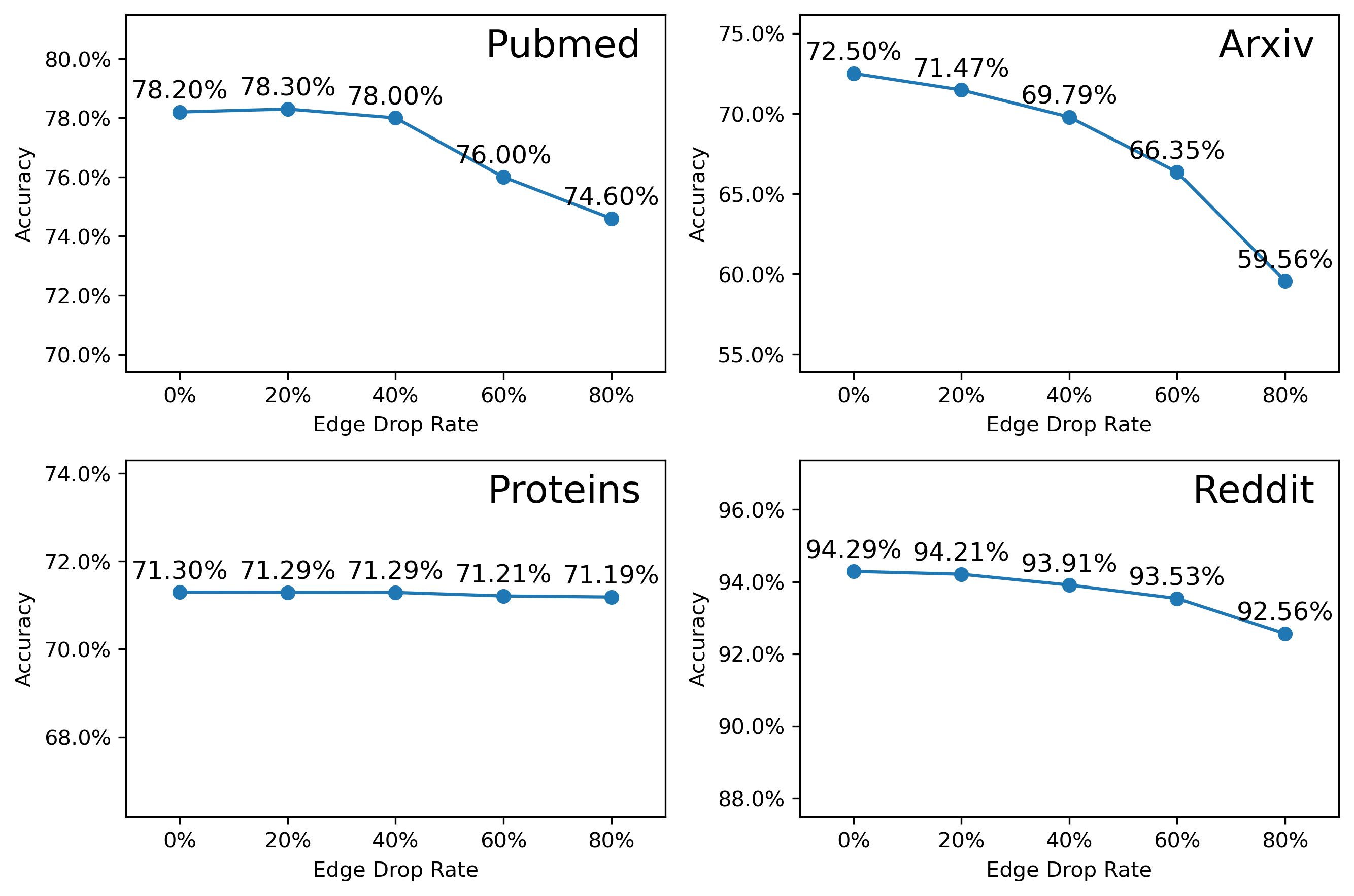}
    \caption{
    GCN's \cite{gcn} inference accuracy when feeding sampled graphs with different edge dropping rates on different datasets. The dropped edges are randomly selected. 
    }
    \label{fig:acc_drop}
\end{figure}

We mimiced DropEdge \cite{DropEdge} by dropping a fixed rate of edges for a given graph, fed the sampled graph to a pre-trained GNN model to obtain test accuracy, and swept the drop rate from 20\% to 80\%.
Figure \ref{fig:acc_drop} shows results with the GCN \cite{gcn} model on four representative datasets. 

On Figure \ref{fig:acc_drop}, we observe that the inference accuracy of the GCN model generally declines as the drop rate increases. 
For small datasets such as Pubmed and Arxiv, accuracy remains close to the baseline when the edge drop rate falls below 40\%.
For large datasets such as Proteins and Reddit, accuracy is less sensitive to larger drop rates compared to smaller datasets, and we observe only a small loss even when the edge drop rate reaches 80\%.

This analysis shows that a pre-trained GNN model can resist the loss of edges, which provides an opportunity to trade off accuracy and latency.
We leverage this fact in our design of \cachesample{}.

\subsection{SpMM Time Reduction and Sampling Time}
\label{subsec:sample_overhead}

As shown in Figure \ref{fig:edge_sample}, edge sampling reduces the NNZ of the sampled graph's adjacency matrix, which can diminish the runtime of the SpMM kernel since less computation is needed.
However, it requires extra compute time to perform edge sampling.
Therefore, an important question becomes whether edge sampling can bring a net end-to-end speedup.

\begin{table}[b!]
\caption{
GCN's \cite{gcn} SpMM kernel time for running inference with the original and edge-sampled graphs, edge sampling time, and the overall inference slowdown after combining the new SpMM kernel time and sampling overhead, for different datasets. 
The edge sampling is implemented in NumPy, and the drop rate is 20\%. 
The time is measured in milliseconds.
}
\label{tab:sample_spmm_time}
\begin{center}
\begin{small}
\begin{sc}
\scalebox{0.9}{
\begin{tabular}{l|rr|c|c|c}
\toprule
\multirow{2}{*}{Dataset} & \multicolumn{2}{c|}{SpMM Time} & SpMM & Sample & Overall \\
& Ori. & Sampled & Speedup & Time & Slowdown \\
\midrule
Pubmed        & 0.68  & 0.65  & 1.05x & 4.24  & 2.1x \\
Arxiv    & 14    & 13.1  & 1.07x & 108.5 & 5.2x \\
Proteins & 173.2 & 149.4 & 1.16x & 13528 & \textbf{75.7x} \\
Reddit        & 99.86 & 89.1  & 1.12x & 6814  & \textbf{63.8x} \\
\bottomrule
\end{tabular}}
\end{sc}
\end{small}
\end{center}
\vskip -0.1in
\end{table}

To answer this question, we conducted another analysis to access how much a SpMM kernel can be accelerated using an edge-sampled graph and how much compute time it takes to perform edge sampling.
We follow the same settings as Section \ref{subsec:bottleneck} and use cuSPARSE's SpMM kernel to conduct this analysis.
We continued to mimic \cite{DropEdge} for sampling methodology.
We implemented edge sampling in NumPy, the standard way used in \cite{DropEdge}, and randomly dropped 20\% of the edges for each graph.
We chose a 20\% drop rate here because the tested model maintained near baseline accuracy with this rate, as Figure \ref{fig:acc_drop} shows.
Table \ref{tab:sample_spmm_time} shows the comparison.

As shown in Table \ref{tab:sample_spmm_time}, the SpMM kernel time is generally faster with the sampled graph, with a speedup range from 1.05x to 1.16x. 
However, when we consider edge sampling time, the net inference time becomes significantly slower due to the huge number of computing cycles spent on the sampling itself. 
Using the \tit{Reddit} dataset as an example, although the SpMM kernel runs 1.12x faster with the sampled graph, edge sampling consumes 6814 ms to run, which results in a 63.8x slower end-to-end inference time.
Here, the used random edge sampling strategy is considered lightweight. 
If we use a more complicated sampling method, such as a neural network model \cite{neural_sparse}, the gap between the sampling time and the SpMM kernel time reduction would be even larger.

As a result, applying existing edge sampling methods on existing SpMM kernels cannot yield a net acceleration to GNN inference, especially with the condition that the graph would be processed only once.
We therefore, propose \cachesample{}, a novel cache-first edge sampling mechanism and SpMM kernel codesign that efficiently leverages edge sampling to significantly accelerate GNN inference.  
\section{CacheSample Kernel Design}
\label{sec:kernel}
We now provide a detailed description of \cachesample{} design, with the goal of leveraging edge sampling to speedup SpMM, the compute bottleneck of GNN inference, while minimizing edge sampling overhead.

\begin{figure}[t!]
    \centering
    \includegraphics[width=1.\linewidth]{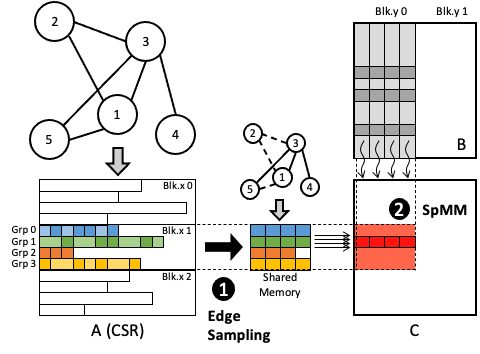}
    \caption{
    Overview of the \cachesample{} kernel. 
    A, a sparse matrix in CSR format, represents the original graph structure. 
    B contains nodes' features, and C contains output features. 
    Both of them are dense matrices in the row-major ordering.
    There are two main stages in \cachesample{}: (1) performing edge sampling and loading the complete edge-sampled graph into shared memory; (2) performing SpMM based on the sampled graph. 
    When loading A's values, a group of threads share the workload of one row of A, which parallelizes the sampling process.
    When loading B's values, threads in the same group load data in the neighboring columns, which creates coalesced memory access.
    Each thread computes the result of each C's element.
    }
    \label{fig:cache_sample}
\end{figure}

\subsection{Overview}
Figure \ref{fig:cache_sample} presents an overview of \cachesample{}'s kernel architecture. 
Input to the kernel is a sparse matrix $A$ in CSR format, which represent the graph structure. 
A dense matrix $B$ contains the nodes' features.
The output of the kernel is a dense matrix C, which contains the output nodes features.
Dense matrices $B$ and $C$ are both in row-major format.

The main computations in \cachesample{} include \tit{edge sampling}, which shrinks the sparse matrix $A$ to fit on the GPU's shared memory, and a \tit{SpMM operation}, compute the output features $C$ based on the sampled graph.
Algorithm \ref{alg:cachesample} lists the pseudo code of \cachesample{}.
We describe each design step in the following subsections.

\begin{algorithm}[t!]
    \caption{Pseudo Code of CacheSample}
    \label{alg:cachesample}
    \begin{algorithmic}[1]
    \STATE {\bfseries Input:} \text{A.data, A.row\_ptr, A.col\_ind, B, shmem\_width}
    \STATE {\bfseries Output:} \text{C}
    \FOR{each thread}
    \STATE \_\_shared\_\_ sh\_data[], sh\_cols[];
    \STATE row\_id, col\_id, sh\_offset, row\_nnz = get\_offsets();
    \STATE S = min(row\_nnz, shmem\_width); 
    \FOR{$i \gets$ sh\_offset \textbf{to} $S$ \textbf{step} $blockDim.x$}
    \STATE sample\_idx = get\_sample\_index($i$, row\_nnz);
    \STATE sh\_data[i], sh\_cols[i] = load\_A(sample\_idx);
    \ENDFOR
    \STATE \_\_syncthreads();
    \STATE acc = 0;
    \FOR{$j \gets 0$ \textbf{to} $S$}
    \STATE acc += sh\_data[j] $\ast$ B[sh\_cols[j], col\_id];
    \ENDFOR
    \STATE C[row\_id, col\_id] = acc;
    \ENDFOR
  \end{algorithmic}
\end{algorithm}
 
\subsection{Stage 1: Perform Cache-First Edge Sampling}
\label{subsec:cache_first_edge_sampling}
To perform edge sampling in \cachesample{}, we first set
a desired shared memory width, $S$.
Then, the size of the shared memory each thread block can request is determined by $(\# \: of \: thread \: groups \times S)$.
Note that this size is limited by the shared memory capacity of the GPU.
The value $S$ also serves as the sampling target for \cachesample{} to downsize the graph data.
As shown in Figure \ref{fig:cache_sample}, for each row of A, at most $S$ non-zero values are fetched to shared memory. 
If $S$ exceeds the NNZ of that row, then the whole row is fetched to shared memory.
In this way, the edge-sampled graph can be fitted to the shared memory within the scope of each thread block.
We refer to this sampling mechanism as "cache-first edge sampling".

The cache-first edge sampling mechanism offers two main benefits:
The first is the reduced computation time required by the SpMM operation in the next stage; 
the second is improved cache locality from keeping all the needed graph data on shared memory.
This is not feasible for existing SpMM kernel implementations because they must ensure the completeness of the SpMM operation, and GPU's shared memory is usually not large enough to hold all graph data without sampling.
However, our proposed cache-first edge sampling scheme exploits the fact that a GNN model can tolerate the loss of edges, and discards redundant edges for better cache locality and faster runtime. 

To perform the proposed cache-first edge sampling efficiently, we assign a group of threads to sample a row of $A$ in \cachesample{}.
Here, each thread uses one of the lightweight sampling strategies (Section \ref{subsec:sampling_strategy}) to compute the sampling indices and loads the selected A's data and column indices into shared memory.
We arrange 32 to 128 threads in each thread group.
Depending on the size of each thread group, we arrange 4 to 8 thread groups in a thread block, forming a fixed 512 threads in each thread block.
As the thread groups in the same thread block share the same shared memory space, each shared memory block holds 4 to 8 rows of $A$'s sampled values.
In the \tit{SpMM} operation, the next stage of \cachesample{}, all values of A come from shared memory, and the kernel computes the SpMM only for the sampled values. 

As noted in Section \ref{subsec:sample_overhead}, edge sampling can cause substantial compute overhead that retards overall runtime.
\cachesample{} therefore, parallelizes the sampling task into thousands of threads and significantly accelerates it.
Such in-kernel sampling also eliminates the need to preprocess graph data on the CPU and lets user-level code remain untouched when using the \cachesample{} as the back-end SpMM kernel. 
In this way, although edge sampling must be re-computed each time the kernel is launched, our evaluation results (Section \ref{sec:eval}) show that \cachesample{} can still yield massive acceleration, which indicates that our parallelized edge sampling design is very efficient.

\subsection{Stage 2: Compute SpMM Based on Sampled Results}
Once edge sampling is done and shared memory is loaded, the second stage of \cachesample{} begins, which aims to compute the SpMM based on sampled results.
Since \cachesample{} computes the multiplications only for the sampled A's values, it approximates the SpMM.

Operations in this stage are straightforward, including reading A's data and column indices from shared memory, loading the corresponding B's data from global memory, calculating the product, and looping to repeat these process and accumulate multiplication results.
Algorithm \ref{alg:cachesample}, lines 12 to 16, shows the pseudo code for this stage.

To perform these operations, \cachesample{} assigns to each thread the computation task for an element of C.
This thread management resembles SpMM kernel optimizations in \cite{merge-spmm} and \cite{gespmm}.
However, the major difference is that all needed A's values in \cachesample{} are loaded to shared memory prior to this stage.
Therefore, the clean and dedicated loop here needs no synchronization or extra data loading, which means that \cachesample{} has fewer branch divergences and can achieve a higher ILP. 

After the sampled SpMM is computed, each thread stores the multiplication result of an element of C from a local register to global memory, and \cachesample{} kernel operations are finished.
The whole \cachesample{} process is implemented in roughly 50 lines of CUDA code.

\begin{figure}[t!]
    \centering
    \includegraphics[width=.95\linewidth]{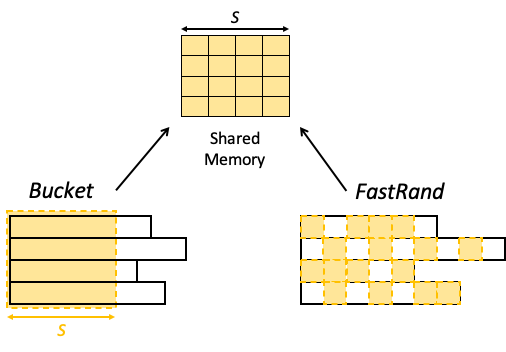}
    \caption{
    Overview of \cachesample{}'s two edge sampling strategies, \buc{} and \fr{}. \buc{} takes the first $S$ continuous A's values. \fr{} samples across the whole region in a pseudo-random manner.
    }
    \label{fig:edge_sampling}
\end{figure}

\subsection{Sampling Strategy}
\label{subsec:sampling_strategy}
Section \ref{subsec:cache_first_edge_sampling} stated that within the cache-first edge sampling mechanism, \cachesample{} could use different lightweight sampling strategies for each thread to compute exact sampling indices (see line 8 in Algorithm \ref{alg:cachesample}). 
We now specify the design of the strategies.

A primary design consideration for \cachesample{}'s sampling is efficiency.
Although we use parallel threads to accelerate edge sampling (Section \ref{subsec:cache_first_edge_sampling}), computation of the sampling itself must still be lightweight to maximize \cachesample{} performance.
Based on this consideration, we designed two lightweight edge sampling strategies, \buc{} and \fr{}, with different speedup and accuracy impacts.
Figure \ref{fig:edge_sampling} highlights each strategy, which we describe below.

\textbf{\buc{}.}
Given the target shared memory width, $S$, the \buc{} edge sampling strategy picks the first $S$ non-zero data and column indices for each row of $A$.
We call this strategy \buc{} because it resembles the way a bucket carries items to fit a specific volume.
\buc{} is very fast since it needs only a boundary check to determine which indices to keep.
However, we found that it could significantly sacrifice accuracy, especially on large graphs because it maintains a fixed range of edges for each node.
See Section \ref{sec:eval} for more information on \buc{}'s accuracy.

\textbf{\fr{}.}
As noted in Section \ref{subsec:edge_sample_acc}, GNN's inference accuracy can be well preserved using a random edge sampling scheme, especially for large graphs with numerous nodes and edges.
Based on this observation, we designed the \fr{} edge sampling strategy to perfor a pseudo-random sampling for \cachesample{}.
To attain both good randomness and fast speed, \fr{} uses an efficient hash function to calculate sample indices;
Equation \ref{eq:fastrand} shows the hash function.

As Equation \ref{eq:fastrand} shows, to calculate the sample index ($sample\_idx$) for a given shared memory location ($shmem\_idx$), \fr{} multiplies $shmem\_idx$ with a prime number $P^{\prime}$ and takes a modulo over the $row\_nnz$, which is the NNZ of the targeted row.
Then, \cachesample{} uses $sample\_idx$ as the index value to fetch A's data and column indices and store them to the shared memory at position $shmem\_idx$ (see line 9 in Algorithm \ref{alg:cachesample}).
Note that in order to sample across the full range of non-zero values for each row of A, the prime number $P^{\prime}$ cannot be too small.
In practice, we chose $P^{\prime} = 577$ when implementing the \cachesample{} kernel.
In this way, \fr{} fulfills the pseudo-random edge sampling using only one multiplication and one modulo operation, which are sufficiently lightweight for each GPU thread to finish within a short latency.
With this type of sampling, we expect the accuracy loss of \fr{} to be less than \buc{}.
However, we expect \buc{} to be faster since it needs less computation.
See Section \ref{sec:eval} for our detailed evaluation.

\begin{equation} 
\label{eq:fastrand}
sample\_idx = (shmem\_idx \times P^{\prime}) \mod row\_nnz
\end{equation}

\subsection{Kernel Optimizations} 
In addition to cache-first edge sampling and the sampled SpMM, \cachesample{} uses other kernel optimizations to achieve optimal performance.
We now describe these.

\subsubsection{Thread management} 
Per Section \ref{subsec:cache_first_edge_sampling}, we use up to 128 threads to handle the sampling and data loading for each row of $A$ in \cachesample{}.
This is unique since a common thread management approach adopted by many existing SpMM kernel optimizations \cite{merge-spmm, gespmm, gale2020sparse, aspt} uses a warp (32) of threads to load a row of A's values to registers or shared memory.
Previous works use this thread setting to provide an acceptable balance of parallelism and number of tasks per thread for SpMM workloads.
However, unlike the previous SpMM kernels, which usually load only a partial row at one time,
\cachesample{} must complete the sampling and data loading for an entire row in one stage. 
Because the targeted shared memory width, $S$, could be large (such as 512), using only 32 threads for a row requires multiple iterations to completely load the shared memory, which may slow down both sampling and data loading.
Therefore, we use up to 128 threads to sample and load a row of $A$ in \cachesample{} to improve performance.

\subsubsection{Coalesced memory access}
Coalesced memory access is an important optimization to reduce memory loading latency on GPU.
In the second stage of \cachesample{}, we allowed a group of threads to load B's data in the same row but in neighboring columns at the same time.
This coalesces memory access because the data requests from multiple threads can be combined and fulfilled by fewer global memory transactions. 
As a result, data loading for B in \cachesample{} is very efficient.
This technique resembles those introduced in \cite{merge-spmm, gespmm}.

\subsubsection{Load balancing}
It is natural for some nodes to have many more connections than other nodes in the graph, leading to a large variations of the NNZ from row to row, as Figure \ref{fig:cache_sample} shows.
The irregular distribution can easily result in an unbalanced load for the SpMM kernel and degrade kernel performance, especially under a row-split work distribution.
In \cachesample{}, the NNZ variation of each row is limited to the targeted shared memory width, $S$, after edge sampling.
In this way, the work assigned to each thread is much better balanced than running on an unsampled graph, helping \cachesample{} achieve higher utilization and improved performance.
\section{Evaluation}
\label{sec:eval}
We next describe system integration, experiment setup, and the comprehensive experiments we conducted to verify the effectiveness of \cachesample{}. 

\subsection{Integration with DGL}
\label{sebsec:integrate_dgl}
To maximize the applicability of \cachesample{} to GNN models, we integrated \cachesample{} into one of the most popular GNN frameworks, DGL \cite{dgl}, which provides a set of high-level Python APIs.
The backend of DGL calls to the cuSPARSE's \textit{cusparseSpMM()} kernel to perform the SpMM operation.

Since both \textit{cusparseSpMM()} and \cachesample{} work on the standard CSR format, we simply replaced the \textit{cusparseSpMM()} with our \cachesample{} kernel in the DGL backend for the integration.
Because \cachesample{} needs no preprocessing on the graph data, the DGL user-level code remains unmodified during this process.

\begin{table}[tb!]
\caption{Model parameters and baseline accuracy we achieve on each dataset for GCN and GraphSage.}
\label{tab:model}
\begin{center}
\begin{small}
\begin{sc}
\scalebox{0.9}{
\begin{tabular}{lccrr}
\toprule
Model & Dataset & \#Layer & \#Hidden & Test Acc \\
\midrule
\multirow{4}{*}{GCN} 
& Pubmed   & 2 & 32    & 77.70\% \\  
& arxiv    & 3 & 256   & 72.50\% \\
& proteins & 3 & 256   & 71.30\% \\
& Reddit   & 2 & 128   & 94.29\% \\
\midrule
\multirow{4}{*}{GraphSage}  
& Pubmed   & 2 & 32  & 78.60\% \\
& arxiv    & 3 & 256 & 72.95\% \\
& proteins & 3 & 256 & 77.28\% \\
& Reddit   & 2 & 128 & 96.22\% \\
\bottomrule
\end{tabular}}
\end{sc}
\end{small}
\end{center}
\vskip -0.1in
\end{table}

\subsection{Experiment Setup}
\label{subsec:exp_setup}
We now describe the setup we used to conduct our experiments. 

\subsubsection{Evaluated GNN models and datasets}
We chose two classic GNN models, \tit{GCN} \cite{gcn} and \tit{GraphSage} \cite{graphsage}, and four node classification datasets to evaluate our \cachesample{} kernel.
The four graph datasets were \tit{Pubmed, Arxiv, Proteins and Reddit}.
\tit{Pubmed} is a citation network of medical academic papers.
\tit{Arxiv} and \tit{Proteins} were taken from the \tit{Open Graph Benchmark - Node Property Prediction} collection \cite{ogb}. 
\tit{Reddit} is a social network dataset collected from a popular online forum.
These four datasets are of different scales in terms of the number of nodes and edge. For \tit{GraphSage}, we used \tit{mean} feature aggregation. 
Hence, experiment results using them can help us appreciate \cachesample{}'s performance for different graph scales.
Table \ref{tab:model} lists the model parameters for each datasets, and Table \ref{tab:dataset} shows the dataset statistics. 

\subsubsection{Model setup}
We first trained the GNN models on each dataset for ten times with the original DGL framework, and kept the ones with the highest test accuracy.
Table \ref{tab:model} shows the best test accuracy we achieved for each model and dataset.
After the models were trained, we used our modified DGL to call \cachesample{} and run the experiments on inference.

\subsubsection{SpMM baselines}
Our experiments compare the performance of \cachesample{} to the following CSR SpMM baselines.
\begin{itemize}
    \item \textit{cuSPARSE}: The default implementation in DGL. The main kernel is \textit{cusparseSpMM()}, and the compute algorithm is set to \textit{CUSPARSE\_SPMM\_CSR\_ALG2} \cite{cusparse_doc}, as it achieves the best performance in our setup.
    \item \textit{Merge-SpMM}: Proposed in \cite{merge-spmm}. It has a suite of different row-split and merge-based algorithms aimed at improving load balance. 
    The code is available at \cite{merge-spmm_code}.
    To test this kernel, we ran it with all proposed algorithms and picked the highest performer in each case. 
    \item \textit{GE-SpMM}: Proposed in \cite{gespmm}. 
    It is based on the row-split design of \cite{merge-spmm} but has an improved shared memory usage.
    We also ran all kernel setup provided in their open-source package \cite{gespmm_code} and picked the best performer.
\end{itemize}

\subsubsection{Hardware environment}
Our experiments were conducted on the following hardware environment.
\begin{itemize}
    \item \textit{AWS EC2 p3.2xlarge instance}. GPU model: NVIDIA V100 with 16GB global memory. CUDA version: 11.0. CPU: Intel Xeon E5-2686 v4 (8 virtual cores) with 61GB main memory.
\end{itemize}

\subsection{Sampling Rate} 
\label{subsec:sample_rate}
As noted in Section \ref{subsec:edge_sample_acc}, sampling rate is an important factor affecting inference accuracy.
In \cachesample{}, the target shared memory width, $S$, determines the sampling rate.
Since each graph has a different distribution of nodes and edges, its sampling rate under each $S$ value is different in \cachesample{}.

To obtain the sampling rates, we simulated the cache-first edge sampling mechanism in Python and calculated the equivalent sampling rates for each $S$.
Table \ref{tab:sample_rate} shows the sampling rates for each dataset.
Note that \buc{} and \fr{} have the same sampling rate for a given $S$ since they fetch the same amount of A's values to shared memory.

Per Table \ref{tab:sample_rate}, for small graphs such as \tit{Pubmed} and \tit{Arxiv}, a small $S$ (16) already holds over 80\% percent of the edges, and a larger $S$ (above 128) actually holds over 99\% of the edges.
However, for large graphs such as \tit{Proteins} and \tit{Reddit}, the sampling rates are much lower compared to the small graphs under a fixed $S$.
For example, when $S = 256$, the sampling rates are only around 34\% for \tit{Proteins} and \tit{Reddit} but are 100\% and 99.9\% for \tit{Pubmed} and \tit{Arxiv}.
Based on our sampling rate analysis, we expect that \cachesample{} will experience little to no accuracy loss for small graphs but a more sizable accuracy loss for large graphs, especially when $S$ is small.

\begin{table}[bt!]
\caption{Sampling rates under different $S$ values for different datasets in \cachesample{}.}
\label{tab:sample_rate}
\begin{center}
\begin{small}
\begin{sc}
\scalebox{0.9}{
\begin{tabular}{lrrrrrr}
\toprule
Dataset & s=16 & s=32 & s=64 & s=128 & s=256 & s=512 \\
\midrule
Pubmed	    & 84.9\%	& 95.8\%	& 99.3\%	& 99.9\%	& 100.0\%	    & 100.0\%	\\
Arxiv	    & 83.7\%	& 96.8\%	& 99.3\%	& 99.8\%	& 99.9\%	& 100.0\%	\\
Proteins	& 2.6\%	& 5.1\%	& 9.9\%	& 18.9\%	& 34.3\%	& 56.7\%	\\
Reddit	    & 3.1\%	& 6.0\%	& 11.3\%	& 20.5\%	& 34.8\%	& 53.9\%	\\
\bottomrule
\end{tabular}}
\end{sc}
\end{small}
\end{center}
\vskip -0.1in
\end{table}

\begin{figure*}[th!]
    \begin{subfigure}{\textwidth}
    \centering
        \includegraphics[width=1\linewidth]{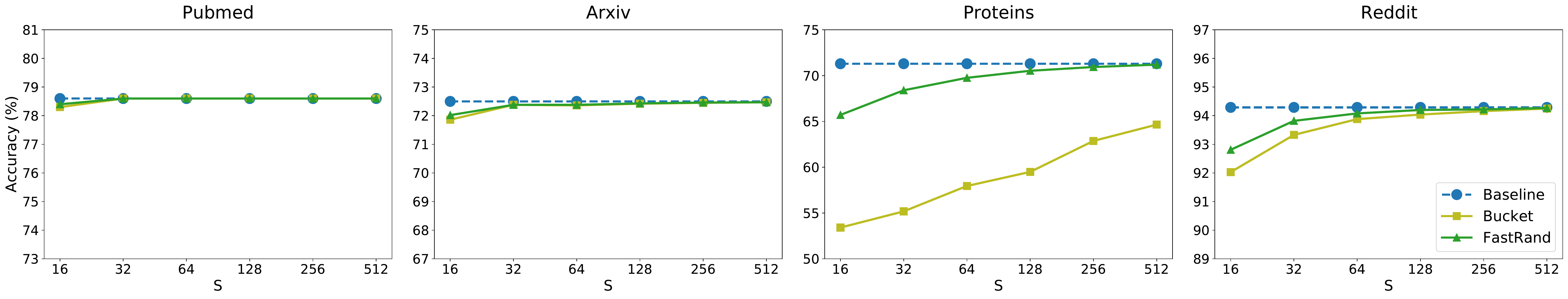}
        \caption{GCN inference accuracy.}
        \label{subfig:cs_acc_gcn}
    \end{subfigure}
    \begin{subfigure}{\textwidth}
        \centering
        \includegraphics[width=1\linewidth]{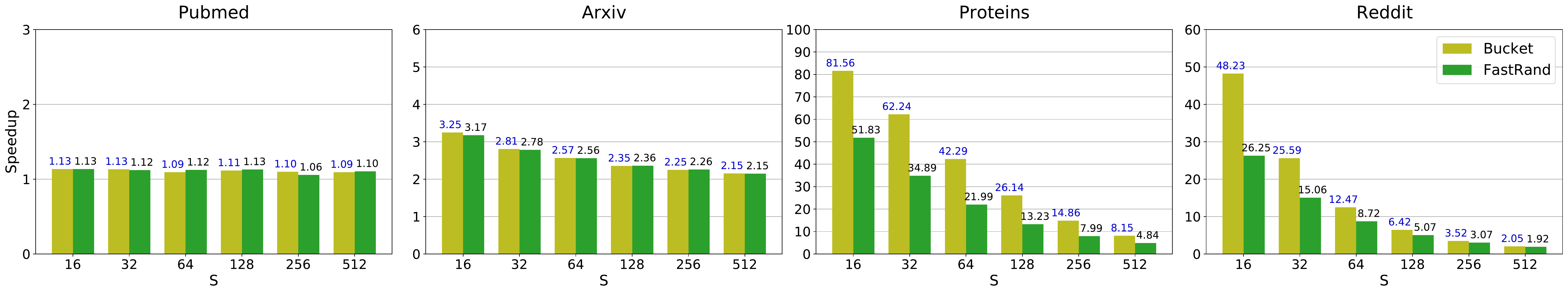}
        \caption{SpMM kernel speedup for \tit{GCN}.}
        \label{subfig:cs_speedup_gcn}
    \end{subfigure}
    \begin{subfigure}{\textwidth}
    \centering
        \includegraphics[width=1\linewidth]{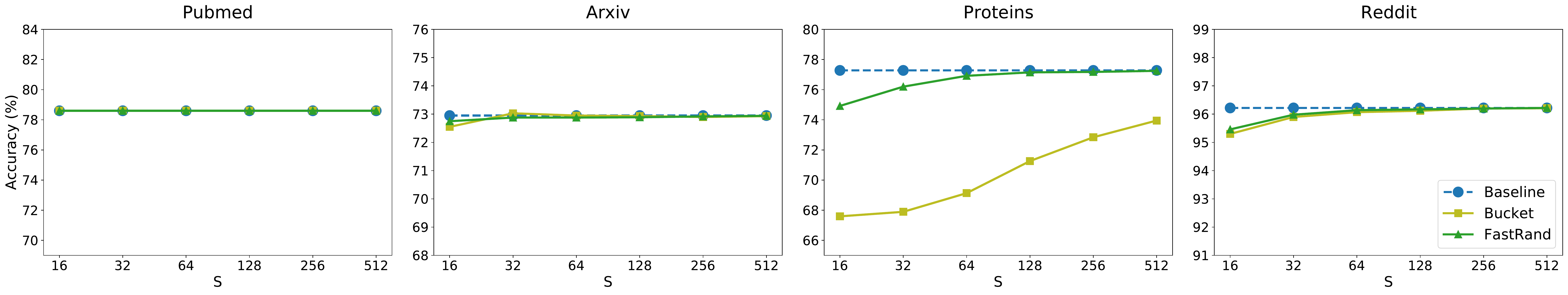}
        \caption{GraphSage inference accuracy.}
        \label{subfig:cs_acc_sage}
    \end{subfigure}
    \begin{subfigure}{\textwidth}
        \centering
        \includegraphics[width=1\linewidth]{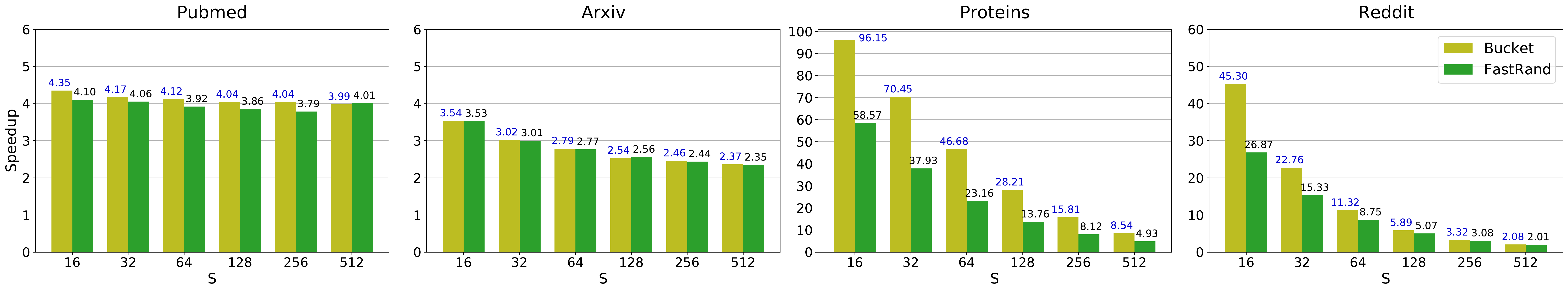}
        \caption{SpMM kernel speedup for \tit{GraphSage}.}
        \label{subfig:cs_speedup_sage}
    \end{subfigure}
    \caption{\cachesample{}'s accuracy and SpMM speedup compared to \tit{cuSPARSE} under different $S$ values (\cachesample{}'s target shared memory width) for \tit{GCN} and \tit{GraphSage}. The experiments were run on NVIDIA V100 GPU.}
    \label{fig:exp_acc_speedup}
\end{figure*}

\subsection{Benefits of \cachesample{}}
\label{subsec:eval_main}
We now present the inference accuracy and speedup of using \cachesample{} compared to cuSPARSE SpMM kernel.

\subsubsection{Inference accuracy and speedup under different $S$ values}
\label{subsubsec:cs_acc_speedup}
To conduct these experiments, we first ran \tit{GCN} and \tit{GraphSage} with the unmodified DGL, which uses cuSPARSE's SpMM kernel, to assess baseline inference accuracy and SpMM kernel time for each dataset.
Here, we used \tit{PyTorch Profiler} \cite{pytorch} to measure SpMM kernel time.
Then, we ran the inference with our modified DGL that uses the \cachesample{} kernel, swept $S$ from 16 to 512, and compared the test accuracy and SpMM performance.
For SpMM performance, we compared the total SpMM kernel time of a complete inference and we took the average after running the inference 10 times.
Figure \ref{fig:exp_acc_speedup} shows experimental results on the accuracy and SpMM speedup against cuSPARSE. 

For accuracy on the small datasets, \tit{Pubmed} and \tit{Arxiv}, 
\cachesample{} generally had little to no accuracy loss across different $S$ values for both \buc{} and \fr{}.
For \tit{GraphSage} on \tit{Pubmed}, \cachesample{} had no accuracy loss even when $S$ was as small as $16$.
This echoes the discussion in Section \ref{subsec:sample_rate} since \cachesample{} can hold most edges with a small $S$ for small graphs.

For SpMM speedup on small datasets, \cachesample{} achieved 1.1x to 4.35x speedup compared to \tit{cuSPARSE}.
For \tit{Pubmed}, \cachesample{} achieved roughly 1.1x and 4x speedup for \tit{GCN} and \tit{GraphSage} respectively, across all $S$ values.
We saw a smaller speedup for \tit{GCN} because the baseline SpMM kernel time was already short (0.64ms). 
The steady speedup across different $S$ values occurred because the sampling rates were at a similar level (Table \ref{tab:sample_rate}).
On \tit{Arxiv}, \cachesample{}'s speedup slightly decreased when $S$ went up.
This is due to increased sampling overhead with a larger $S$, which diminished \cachesample{}'s performance.
However, even with a large $S = 512$, \cachesample{} still achieved at least a 2.15x and 2.35x speedup for \tit{GCN} and \tit{GraphSage} respectively, on \tit{Arxiv}.

Regarding accuracy on the datasets \tit{Proteins} and \tit{Reddit}, \cachesample{} experienced a larger loss of accuracy but still achieved near-baseline accuracy with a larger $S$ value.
Also, on these larger graphs, \fr{} generally exhibited better accuracy than \buc{}.
For example, on \tit{Proteins}, \buc{} had a large accuracy loss even when $S = 512$.
In our experiments, we found that $S$ must be as large as $1792$ to achieve less than a 1\% accuracy loss when using \buc{} on \tit{Proteins}.
However, for \fr{}, it achieved near-baseline accuracy with $S = 128$ for \tit{GCN} and $S = 64$ for \tit{GraphSage}.
\fr{} also had much lower drop in accuracy than \buc{} when $S$ was smaller;
this is because the graph degree of \tit{Proteins} is large, 
and the \buc{}'s method for picking the first continuous edges makes it lose many graph features.
On the other hand, \fr{} randomly sampled all edges, which preserved graph structure and yield much improved accuracy.
For \tit{Reddit}, we observed an accuracy trend similar to \tit{Proteins}.
However, since \tit{Reddit} is an easier dataset that both \tit{GCN} and \tit{GraphSage} achieve high baseline accuracy (94.3\% and 96.2\%), \buc{}'s accuracy loss on \tit{Reddit} is not as pronounced as \tit{Proteins}.

Regarding SpMM speedup on large datasets, \cachesample{} achieved a tremendous speedup agains \tit{cuSPARSE} for both \buc{} and \fr{}, due to the low effective sample rates. 
On \tit{Proteins}, although \cachesample{} with \buc{} achieved up to 81.6x and 96.15x speedup for \tit{GCN} and \tit{GraphSage} respectively, it was less meaningful since the accuracy loss was large.
However, with \fr{}, \cachesample{} achieved a 13x speedup with only a negligible accuracy loss when $S = 128$ for both \tit{GCN} and \tit{GraphSage} on \tit{Proteins}.
On \tit{Reddit}, both \buc{} and \fr{} achieved a meaningful speedup with only a small accuracy loss.
When $S = 64$, \buc{} achieved 12.5x and 11.3x and \fr{} achieved 8.72x and 8.75x for \tit{GCN} and \tit{GraphSage} respectively, where both strategies showed negligible accuracy loss, a significant benefit due to \cachesample{}.


\begin{scriptsize}
\begin{table*}[tbh!]
    \centering
    \caption{\cachesample{}'s best SpMM speedup against cuSPARSE with less than 1\% accuracy loss.}
    \label{tab:best}
    \scalebox{0.925}{
    \begin{tabular}{lc|rr|rrr|rr|rrr|rr}
        \toprule
        \multirow{2}{*}{Model} & \multirow{2}{*}{Dataset} & \multicolumn{2}{c|}{$S$} &  \multicolumn{3}{c|}{Accuracy} & \multicolumn{2}{c|}{Accuracy Loss} & \multicolumn{3}{c|}{SpMM Time (ms)} & \multicolumn{2}{c}{SpMM Speedup} \\
        & & \buc{} & \fr{} & Baseline & \buc{} & \fr{} & \buc{} & \fr{} & Baseline & \buc{} & \fr{} &  \buc{} & \fr{} \\
        \midrule
        \multirow{4}{*}{GCN}	
        & Pubmed    & 16 & 16 & 77.70\%	& 77.30\% & 77.50\%	& 0.40\% & 0.20\% & 0.64 & 0.56 & 0.56 & 1.13 & 1.13 \\
        & Arxiv	    & 16 & 16 & 72.50\%	& 71.86\% & 72.02\%	& 0.64\% & 0.48\% & 12.39 & 3.81 & 3.91	& 3.25 & 3.17 \\
        & Proteins	& 1792	& 128 & 71.30\%	& 70.72\% & 70.53\% & 0.58\% & 0.77\% & 172.92	& 71.93 & 13.07	& 2.40	& \textbf{13.23} \\
        & Reddit	& 32 & 32 & 94.29\%	& 93.33\% & 93.82\%	& 0.96\% & 0.47\% & 99.74 &3.90	& 6.62 & \textbf{25.59} & \textbf{15.06} \\
        \midrule
        \multirow{4}{*}{GraphSage}
        & Pubmed	&16	& 16 & 78.60\% & 78.60\% & 78.60\% & \textbf{0.00\%}	& \textbf{0.00\%} & 3.56	& 0.82 & 0.87 & 4.35 & 4.10 \\
        & Arxiv	    & 16 & 16 & 72.95\%	& 72.55\% & 72.75\%	& 0.40\% & 0.20\% & 18.83 & 5.32 & 5.33	& 3.54 & 3.53 \\
        & Proteins	& 2304 & 64 & 77.28\% & 76.47\% & 76.92\% & 0.81\% & 0.36\%	& 237.50 & 140.02 & 10.25 & 1.70 & \textbf{23.16} \\
        & Reddit	& 16 & 16 & 96.22\%	& 95.30\% & 95.46\% & 0.92\% & 0.76\% & 449.69 & 9.93 & 16.74 & \textbf{45.30} & \textbf{26.87} \\
        \bottomrule
    \end{tabular}
    }
    \vskip -0.1in
\end{table*}
\end{scriptsize}

\subsubsection{Speedup with less than 1\% accuracy loss}
Here, we set a 1\% accuracy loss constraint and report the best \cachesample{} SpMM speedup, the corresponding $S$ value, and inference accuracy for each sampling strategy in Table \ref{tab:best}; we still compare to cuSPARSE for speedup.
We set this constraint because we assume that a 1\% accuracy loss is an acceptable cost to trade off for a faster GNN inference time.
In other words, if users can accept a minimally higher accuracy loss, then \cachesample{} can deliver a more significant speedup.

As Table \ref{tab:best} shows, for most of cases, $S = 16$ was sufficient for \cachesample{} to meet this constraint.
For the small datasets, \cachesample{} achieved up to 3.25x for \tit{GCN} and 4.35x for \tit{GraphSage}, both using \buc{}.
For the large datasets, \cachesample{} generally offered an over 10x speedup against the strong cuSPARSE baseline, which is significant.
The best SpMM speedup \cachesample{} achieved for \tit{Proteins} and \tit{Reddit} under this constraint were 23.16x with \fr{} and 45.3x with \buc{}.
The only exception was that \cachesample{} needed a large $S$ to meet this constraint when using \buc{} on \tit{Proteins}.
However, \cachesample{} with \buc{} still achieved a profitable 2.4x and 1.7x speedup for \tit{GCN} and \tit{GraphSage} respectively, with $S = 1792$ and $S = 2304$.

\begin{figure}[tb!]
    \centering
    \includegraphics[width=\linewidth]{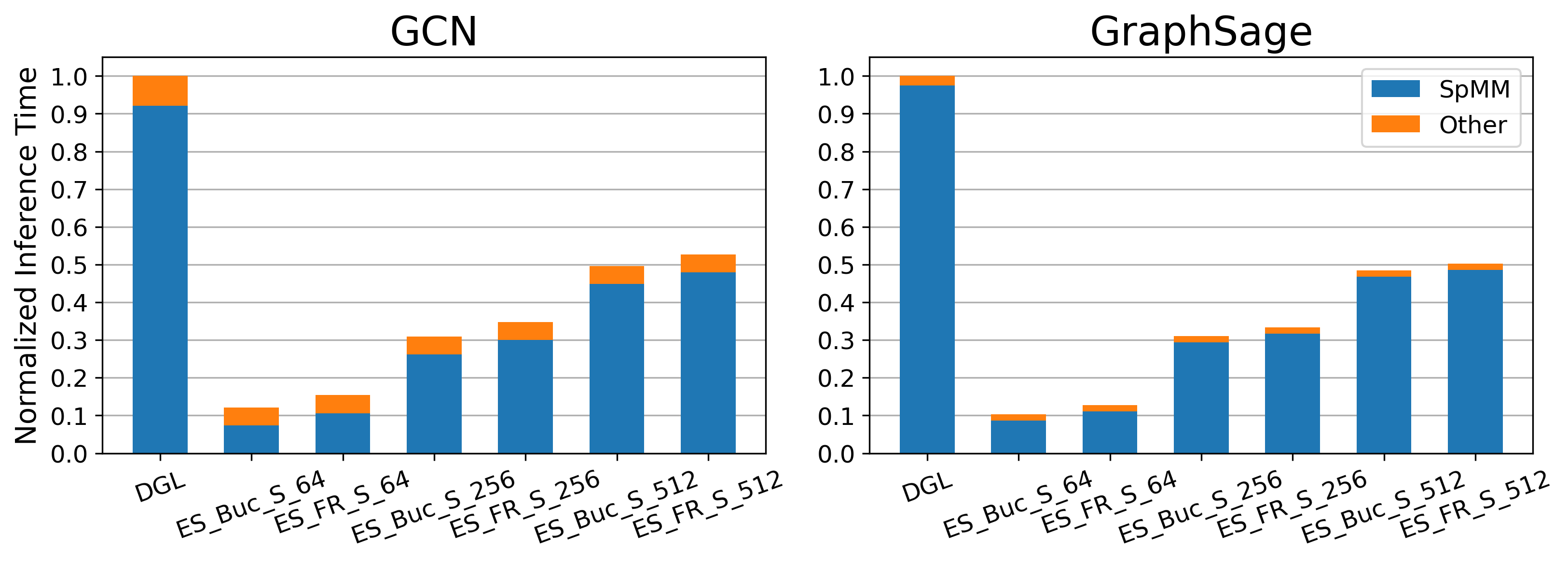}
    \caption{End-to-end inference time breakdown of \tit{GCN} and \tit{GraphSage} on \tit{Reddit}, using the original DGL (calls to \tit{cuSPARSE}) and \cachesample{} with different $S$ values. 
    The time is normalized to DGL's result for each model respectively.}
    \label{fig:cs_e2e}
\end{figure}

\subsubsection{End-to-end inference time} 
Here we compare the end-to-end inference time of using \cachesample{} to the original DGL's results, which uses \tit{cuSPARSE} for its SpMM operation.
Figure \ref{fig:cs_e2e} shows the end-to-end inference time breakdown of running \tit{GCN} and \tit{GraphSage} on \tit{Reddit} with the original DGL and with the \cachesample{} in different $S$ values.
As noted in Section \ref{subsec:bottleneck}: SpMM consumes a large portion of GNN inference time, and noted in Section \ref{subsubsec:cs_acc_speedup}: \cachesample{} significantly accelerates the SpMM operation, the end-to-end inference time was considerably reduced when using \cachesample{} as Figure \ref{fig:cs_e2e} shows.
\cachesample{} achieves about 10x faster end-to-end inference time for both models compared to the original DGL, when using a small $S$ (64).
When using a large $S$ (512), \cachesample{} still achieves roughly 2x speedup.


\subsection{SpMM Performance Compared to Open-Source Baselines}
Besides \tit{cuSPARSE}, we compared \cachesample{}'s SpMM performance to \tit{Merge-SpMM}'s and \tit{GE-SpMM}'s.

To conduct these evaluations, we did not modify the open-source packages of \tit{Merge-SpMM} and \tit{GE-SpMM}, and fed the SpMM kernels with the adjacency matrix extracted from DGL.
This methodology prevented us from changing kernel behavior while re-implementing the kernels into DGL's backend.
To get a similar comparison to the one we achieved in Section \ref{subsec:eval_main}, we separately performed multiple SpMM operations that were needed for a complete inference, and combined the kernel times to get total SpMM time.
We also did the same operations on \tit{cuSPARSE} and \cachesample{} for a fair evaluation.
The results we show are the average of 200 runs.

\begin{figure*}[tbh!]
    \centering
    \includegraphics[width=0.99\linewidth]{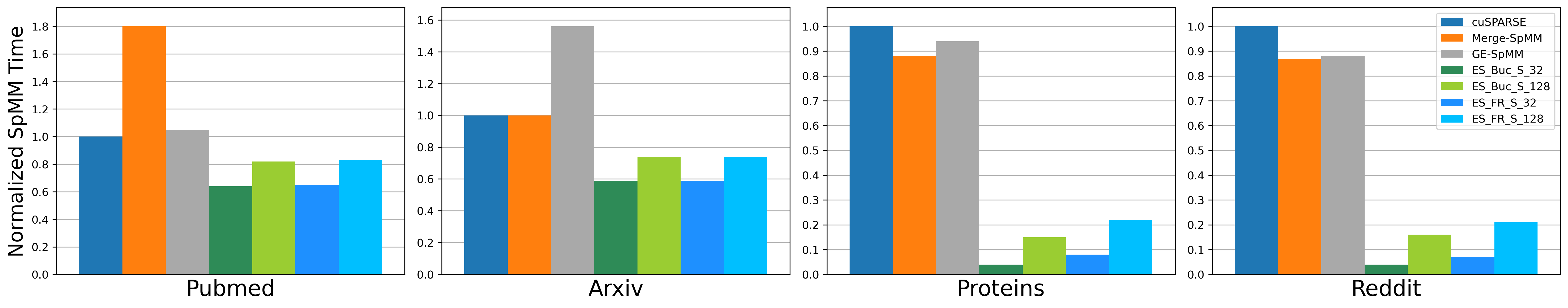}
    \caption{SpMM time of GCN's inference using different SpMM kernels. The time is normalized to cuSPARSE's result for each dataset respectively.}
    \label{fig:gcn_speedup_baselines}
\end{figure*}

Figure \ref{fig:gcn_speedup_baselines} shows the SpMM time comparison of running GCN inference on each dataset.
For \cachesample{}, we show the results of using $S = 32$ and $S = 128$ since these settings were generally optimal for \cachesample{} to achieve near-baseline accuracy for small and large graphs.
As Figure \ref{fig:gcn_speedup_baselines} shows, \tit{Merge-SpMM} and \tit{GE-SpMM} only achieved a slightly faster kernel time than \tit{cuSPARSE} on large graphs. 
However, \cachesample{} achieved the fastest SpMM kernel time across all datasets and excelled on the large graphs, 
consuming only 5\% to 20\% of \tit{cuSPARSE}'s kernel time.

\subsection{Feeding Pre-Sampled Graphs to Baseline SpMM Kernels}
To evaluate the source of \cachesample{}'s benefits, we reran an experiment (Section \ref{subsec:sample_overhead}) that fed pre-sampled graphs to the SpMM kernels.
Unlike Section \ref{subsec:sample_overhead} experiment doing a general random sampling, here we simulated the \fr{} of \cachesample{} in Python for pre-sampling.
Then, we fed the pre-sampled graph to \tit{cuSPARSE}, \tit{Merge-SpMM} and \tit{GE-SpMM} and measured the SpMM kernel time. 
For \cachesample{}, we used the normal graph but set $S$ to the corresponding value. 
Again, we summed SpMM kernel times for a complete inference and took the average of 200 runs for comparison.

\begin{figure}[tb!]
    \centering
    \includegraphics[width=\linewidth]{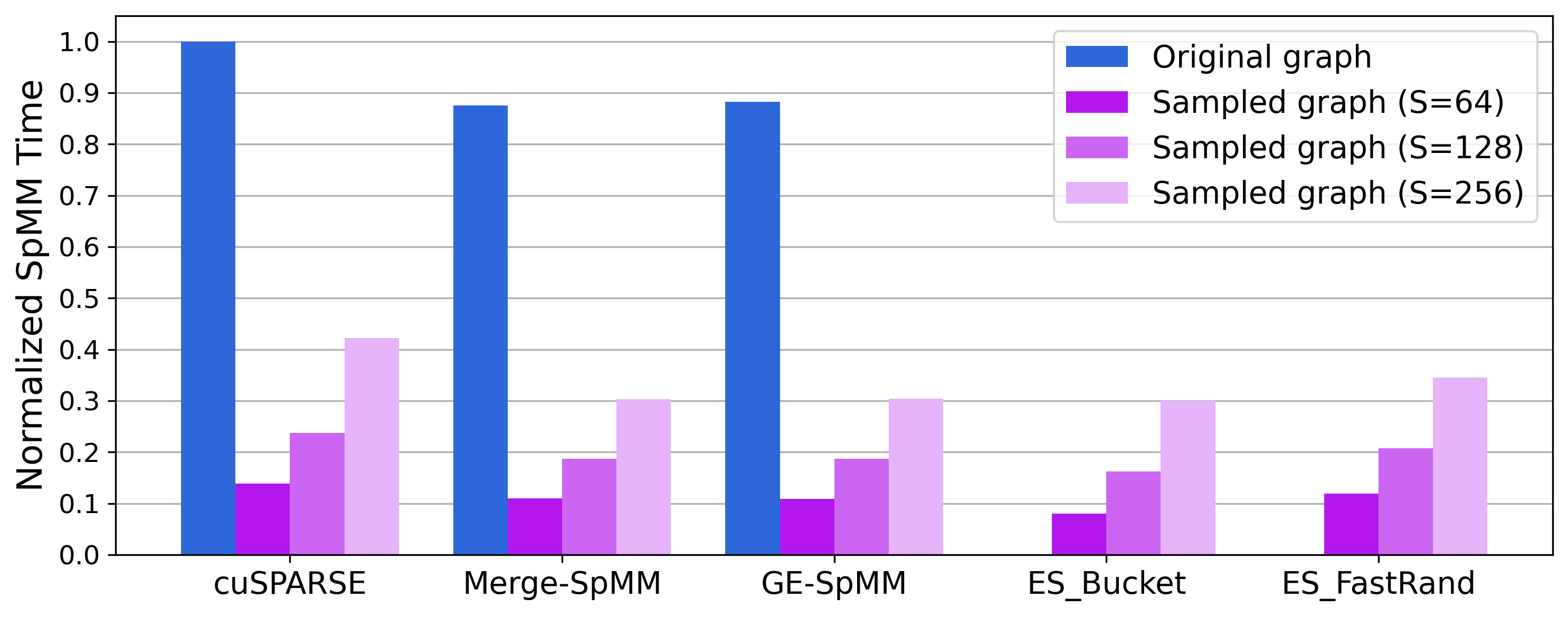}
    \caption{SpMM time of GCN's inference on \tit{Reddit} with the original and pre-sampled graphs. The time is normalized to the result of cuSPARSE with the original graph.}
    \label{fig:gcn_sampled_reddit}
\end{figure}

Figure \ref{fig:gcn_sampled_reddit} shows the SpMM time comparison of running \tit{GCN} on original and pre-sampled \tit{Reddit} with $S = 128$ and $256$.
As the figure shows, with pre-sampled graphs, all baseline kernels had a dramatically faster kernel time. 
For \tit{Merge-SpMM} and \tit{GE-SpMM}, they actually achieved a similar to slightly better performance than \cachesample{} with pre-sampled graphs.
\tit{cuSPARSE} achieved a slightly slower SpMM time than \cachesample{}, but the results were close.
These findings show that the major benefits of \cachesample{} come from the computation reduction enabled by edge sampling.
However, additional optimizations (Section \ref{sec:kernel}) helped \cachesample{} achieve a competitive performance with other highly optimized SpMM kernels when the input graph was already edge-sampled.
Note that \cachesample{} needed to perform edge sampling in the kernel, which took extra overhead that other SpMM kernels did not have.
\cachesample{}'s kernel time can be shortened if we remove time spent on sampling.
We didn't show the results here because such in-kernel time breakdown is difficult to measure.

\section{Discussion and Future Work}
\label{sec:discussion}

\subsection{Normalization of GNN}
A GNN using sum aggregation, generally it normalizes node features by dividing them with the degree of each node.
Practically, a GNN framework like DGL uses a separate operator to handle such normalization.
However, this does not work for \cachesample{} since it reduces the number of edges internally in the SpMM kernel without modifying the graph managed by the framework.
To make the normalization work, we must remove normalization from the DGL and compute it in the \cachesample{} kernel.
Although gaining overhead to compute the normalization, \cachesample{} still achieves a massive speedup and considerable end-to-end acceleration for GNN inference, as Section \ref{sec:eval} shows.

\subsection{Accelerate GNN training}
Although we focus on accelerating GNN inference in this work, \cachesample{} can be used to accelerate GNN training, which also involves many SpMM operations.
However, with our current designs, \cachesample{} would experience lower accuracy when applied to GNN training.
Unlike the inference scheme, users are willing to pay more compute time to improve accuracy during training.
Therefore, applying \cachesample{} to accelerate GNN training does not make sense.

\cachesample{} reduces training accuracy because it samples the same subset of edges each time for a given graph, sampling strategy (\buc{} or \fr{}) and $S$ value. 
As DropEdge \cite{DropEdge} suggests, training accuracy can be maintained or improved by taking a different subset of edges throughout training iterations.
We expect \cachesample{} could accelerate GNN training without losing accuracy if we would efficiently implement a dynamic random sampling into the \cachesample{} kernel.
We leave this to future work.

\subsection{Edge Sampling vs. Node Sampling}
Besides the edge sampling used in this work, there is another kind of sampling, \tit{node sampling}, that is used in GNNs.
Here we briefly introduce the node sampling, clarify the differences between edge sampling, and discuss \cachesample{}'s applicability to it.

Node sampling is first proposed by \cite{graphsage}.
Unlike edge sampling which only takes subsets of edges, node sampling not only takes subsets of nodes but also discards unrelated edges for each subset.
The main usage of existing node sampling methods \cite{graphsage, fastgcn, cluster-gcn, graphsaint, as-gcn} is to extract subgraphs and form the mini-batch training scheme for GNNs, allowing training of very large graphs that cannot be fitted into a single GPU's global memory.

Although the compute time of each node-sampled graph is much shorter because fewer nodes and edges are presented, it requires multiple iterations to run through the original graph.
Therefore, the overall runtime to train or infer for a full graph is increased with node sampling since iterations over different subgraphs incur extra overhead \cite{graphsage, fastgcn}; sometimes, node sampling could even cause high traffics between CPU's and GPU's memory \cite{pagragh, pytorch-direct}, which would significantly slowdown both training and inference.
We expect \cachesample{} to further improve the inference speed for node-sampled graphs by applying edge sampling.
However, the bottleneck will likely be the looping overhead or memory bandwidth contention and thus we expect limited speedup.
\section{Conclusion}
\label{sec:conclusion}

This work proposed \cachesample{}, a cache-first edge sampling mechanism and SpMM kernel codesign.
The \cachesample{} design leverages the fact that a GNN model can tolerate the loss of edges without losing much accuracy for inference.
By strategically and efficiently sampling the edges and fitting all sampled edges on GPU's shared memory, \cachesample{} can significantly accelerate the SpMM operation with minimum accuracy loss compared to other state-of-the-art SpMM kernels. 
Our experimental results on representative GNN models and datasets
show that \cachesample{} outperforms \tit{cuSPARSE} by up to 4.35x with no accuracy loss and by 45.3x with less than a 1\% accuracy loss.
To the best of our knowledge, ours is the first work to utilize edge sampling to accelerate GNN inference.




\bibliographystyle{IEEEtran}
\bibliography{references}

\begin{thebibliography}{10}
\providecommand{\url}[1]{#1}
\csname url@samestyle\endcsname
\providecommand{\newblock}{\relax}
\providecommand{\bibinfo}[2]{#2}
\providecommand{\BIBentrySTDinterwordspacing}{\spaceskip=0pt\relax}
\providecommand{\BIBentryALTinterwordstretchfactor}{4}
\providecommand{\BIBentryALTinterwordspacing}{\spaceskip=\fontdimen2\font plus
\BIBentryALTinterwordstretchfactor\fontdimen3\font minus
  \fontdimen4\font\relax}
\providecommand{\BIBforeignlanguage}[2]{{%
\expandafter\ifx\csname l@#1\endcsname\relax
\typeout{** WARNING: IEEEtran.bst: No hyphenation pattern has been}%
\typeout{** loaded for the language `#1'. Using the pattern for}%
\typeout{** the default language instead.}%
\else
\language=\csname l@#1\endcsname
\fi
#2}}
\providecommand{\BIBdecl}{\relax}
\BIBdecl

\bibitem{gcn}
T.~N. Kipf and M.~Welling, ``Semi-supervised classification with graph
  convolutional networks,'' in \emph{International Conference on Learning
  Representations (ICLR)}, 2017.

\bibitem{graphsage}
W.~L. Hamilton, R.~Ying, and J.~Leskovec, ``Inductive representation learning
  on large graphs,'' in \emph{NIPS}, 2017.

\bibitem{gnnapp_program}
M.~Allamanis, M.~Brockschmidt, and M.~Khademi, ``Learning to represent programs
  with graphs,'' in \emph{International Conference on Learning
  Representations}, 2018.

\bibitem{gnnapp_physics}
\BIBentryALTinterwordspacing
P.~Battaglia, J.~B.~C. Hamrick, V.~Bapst, A.~Sanchez, V.~Zambaldi,
  M.~Malinowski, A.~Tacchetti, D.~Raposo, A.~Santoro, R.~Faulkner, C.~Gulcehre,
  F.~Song, A.~Ballard, J.~Gilmer, G.~E. Dahl, A.~Vaswani, K.~Allen, C.~Nash,
  V.~J. Langston, C.~Dyer, N.~Heess, D.~Wierstra, P.~Kohli, M.~Botvinick,
  O.~Vinyals, Y.~Li, and R.~Pascanu, ``Relational inductive biases, deep
  learning, and graph networks,'' \emph{arXiv}, 2018. [Online]. Available:
  \url{https://arxiv.org/pdf/1806.01261.pdf}
\BIBentrySTDinterwordspacing

\bibitem{gnnapp_drug}
M.~Zitnik, M.~Agrawal, and J.~Leskovec, ``Modeling polypharmacy side effects
  with graph convolutional networks,'' \emph{Bioinformatics}, vol.~34, no.~13,
  p. 457–466, 2018.

\bibitem{featgraph}
Y.~Hu, Z.~Ye, M.~Wang, J.~Yu, D.~Zheng, M.~Li, Z.~Zhang, Z.~Zhang, and Y.~Wang,
  ``Featgraph: A flexible and efficient backend for graph neural network
  systems,'' in \emph{International Conference for High Performance Computing,
  Networking, Storage, and Analysis (SC)}, 2020.

\bibitem{gespmm}
G.~Huang, G.~Dai, Y.~Wang, and H.~Yang, ``Ge-spmm: General-purpose sparse
  matrix-matrix multiplication on gpus for graph neural networks,'' 2020.

\bibitem{merge-spmm}
C.~Yang, A.~Buluc, and J.~D. Owens, ``Design principles for sparse matrix
  multiplication on the gpu,'' in \emph{European Conference on Parallel
  Processing}, 2018.

\bibitem{ellpack_spmm}
G.~{Ortega}, F.~{Vázquez}, I.~{García}, and E.~M. {Garzón}, ``Fastspmm: An
  efficient library for sparse matrix matrix product on gpus,'' \emph{The
  Computer Journal}, 2014.

\bibitem{block_spmm}
S.~Gray, A.~Radford, and D.~P. Kingma, ``Gpu kernels for block-sparse
  weights,'' \emph{arXiv preprint arXiv:1711.09224}, 2017.

\bibitem{aspt}
C.~Hong, A.~Sukumaran-Rajam, I.~Nisa, K.~Singh, and P.~Sadayappan, ``Adaptive
  sparse tiling for sparse matrix multiplication,'' in \emph{Proceedings of the
  24th Symposium on Principles and Practice of Parallel Programming}, ser.
  PPoPP '19, 2019, p. 300–314.

\bibitem{gale2020sparse}
T.~Gale, M.~Zaharia, C.~Young, and E.~Elsen, ``Sparse gpu kernels for deep
  learning,'' 2020.

\bibitem{DropEdge}
\BIBentryALTinterwordspacing
Y.~Rong, W.~Huang, T.~Xu, and J.~Huang, ``Dropedge: Towards deep graph
  convolutional networks on node classification,'' in \emph{International
  Conference on Learning Representations}, 2020. [Online]. Available:
  \url{https://openreview.net/forum?id=Hkx1qkrKPr}
\BIBentrySTDinterwordspacing

\bibitem{bayesian_edge_sample}
A.~Hasanzadeh, E.~Hajiramezanali, S.~Boluki, M.~Zhou, N.~Duffield,
  K.~Narayanan, and X.~Qian, ``{B}ayesian graph neural networks with adaptive
  connection sampling,'' in \emph{Proceedings of the 37th International
  Conference on Machine Learning}, ser. Proceedings of Machine Learning
  Research, vol. 119.\hskip 1em plus 0.5em minus 0.4em\relax PMLR, 2020, pp.
  4094--4104.

\bibitem{neural_sparse}
C.~Zheng, B.~Zong, W.~Cheng, D.~Song, J.~Ni, W.~Yu, H.~Chen, and W.~Wang,
  ``Robust graph representation learning via neural sparsification,'' in
  \emph{Proceedings of the 37th International Conference on Machine Learning},
  ser. Proceedings of Machine Learning Research, 2020.

\bibitem{li2018}
Q.~Li, Z.~Han, and X.-M. Wu, ``Deeper insights into graph convolutional
  networks for semi-supervised learning,'' in \emph{Thirty-Second AAAI
  Conference on Artificial Intelligence}, 2018.

\bibitem{jknet}
K.~Xu, C.~Li, Y.~Tian, T.~Sonobe, K.~ichi Kawarabayashi, and S.~Jegelka,
  ``Representation learning on graphs with jumping knowledge networks,'' 2018.

\bibitem{klicpera2019}
J.~Klicpera, A.~Bojchevski, and S.~Gunneman, ``Predict then propagate: Graph
  neural networks meet personalized pagerank,'' in \emph{Proceedings of the 7th
  International Conference on Learning Representations}, 2019.

\bibitem{fastgcn}
\BIBentryALTinterwordspacing
J.~Chen, T.~Ma, and C.~Xiao, ``Fast{GCN}: Fast learning with graph
  convolutional networks via importance sampling,'' in \emph{International
  Conference on Learning Representations}, 2018. [Online]. Available:
  \url{https://openreview.net/forum?id=rytstxWAW}
\BIBentrySTDinterwordspacing

\bibitem{cluster-gcn}
\BIBentryALTinterwordspacing
W.-L. Chiang, X.~Liu, S.~Si, Y.~Li, S.~Bengio, and C.-J. Hsieh, ``Cluster-gcn:
  An efficient algorithm for training deep and large graph convolutional
  networks,'' in \emph{Proceedings of the 25th ACM SIGKDD International
  Conference on Knowledge Discovery; Data Mining}, ser. KDD '19, 2019.
  [Online]. Available: \url{https://doi.org/10.1145/3292500.3330925}
\BIBentrySTDinterwordspacing

\bibitem{as-gcn}
W.~Huang, T.~Zhang, Y.~Rong, and J.~Huang, ``Adaptive sampling towards fast
  graph representation learning,'' in \emph{Advances in Neural Information
  Processing Systems (NIPS)}, 2018.

\bibitem{graphsaint}
\BIBentryALTinterwordspacing
H.~Zeng, H.~Zhou, A.~Srivastava, R.~Kannan, and V.~Prasanna, ``{GraphSAINT}:
  Graph sampling based inductive learning method,'' in \emph{International
  Conference on Learning Representations}, 2020. [Online]. Available:
  \url{https://openreview.net/forum?id=BJe8pkHFwS}
\BIBentrySTDinterwordspacing

\bibitem{deepgcns}
G.~Li, M.~Müller, A.~Thabet, and B.~Ghanem, ``Deepgcns: Can gcns go as deep as
  cnns?'' in \emph{The IEEE International Conference on Computer Vision
  (ICCV)}, 2019.

\bibitem{gcnii}
Z.~W. Ming~Chen, B.~D. Zengfeng~Huang, and Y.~Li, ``Simple and deep graph
  convolutional networks,'' 2020.

\bibitem{ogb}
W.~Hu, M.~Fey, M.~Zitnik, Y.~Dong, H.~Ren, B.~Liu, M.~Catasta, and J.~Leskovec,
  ``Open graph benchmark: Datasets for machine learning on graphs,''
  \emph{arXiv preprint arXiv:2005.00687}, 2020.

\bibitem{dgl}
\BIBentryALTinterwordspacing
M.~Wang, L.~Yu, D.~Zheng, Q.~Gan, Y.~Gai, Z.~Ye, M.~Li, J.~Zhou, Q.~Huang,
  C.~Ma, Z.~Huang, Q.~Guo, H.~Zhang, H.~Lin, J.~Zhao, J.~Li, A.~J. Smola, and
  Z.~Zhang, ``Deep graph library: Towards efficient and scalable deep learning
  on graphs,'' 2019. [Online]. Available: \url{http://arxiv.org/abs/1909.01315}
\BIBentrySTDinterwordspacing

\bibitem{cusparse_doc}
\BIBentryALTinterwordspacing
NVIDIA, ``The api reference guide for cusparse, the cuda sparse matrix
  library,'' 2021. [Online]. Available:
  \url{https://docs.nvidia.com/cuda/cusparse/index.html}
\BIBentrySTDinterwordspacing

\bibitem{merge-spmm_code}
\BIBentryALTinterwordspacing
C.~Yang and miheer vaidya, ``Merge-spmm open source code,'' 2020. [Online].
  Available: \url{https://github.com/owensgroup/merge-spmm}
\BIBentrySTDinterwordspacing

\bibitem{gespmm_code}
\BIBentryALTinterwordspacing
G.~Huang, ``Ge-spmm open source code,'' 2020. [Online]. Available:
  \url{https://github.com/hgyhungry/ge-spmm}
\BIBentrySTDinterwordspacing

\bibitem{pytorch}
A.~Paszke, S.~Gross, F.~Massa, A.~Lerer, J.~Bradbury, G.~Chanan, T.~Killeen,
  Z.~Lin, N.~Gimelshein, L.~Antiga, A.~Desmaison, A.~Kopf, E.~Yang, Z.~DeVito,
  M.~Raison, A.~Tejani, S.~Chilamkurthy, B.~Steiner, L.~Fang, J.~Bai, and
  S.~Chintala, ``Pytorch: An imperative style, high-performance deep learning
  library,'' in \emph{Advances in Neural Information Processing Systems 32}.

\bibitem{pagragh}
Z.~Lin, C.~Li, Y.~Miao, Y.~Liu, and Y.~Xu, ``Pagraph: Scaling gnn training on
  large graphs via computation-aware caching,'' in \emph{Proceedings of the
  11th ACM Symposium on Cloud Computing}, ser. SoCC '20.\hskip 1em plus 0.5em
  minus 0.4em\relax New York, NY, USA: Association for Computing Machinery,
  2020.

\bibitem{pytorch-direct}
S.~W. Min, K.~Wu, S.~Huang, M.~Hidayetoğlu, J.~Xiong, E.~Ebrahimi, D.~Chen,
  and W.~mei Hwu, ``Large graph convolutional network training with
  gpu-oriented data communication architecture,'' 2021.

\end{thebibliography}

\end{document}